\documentclass{article}

\usepackage{PRIMEarxiv}

\usepackage[utf8]{inputenc} 
\usepackage[T1]{fontenc}    
\usepackage{hyperref}       
\usepackage{url}            
\usepackage{booktabs}       
\usepackage{amsfonts}       
\usepackage{nicefrac}       
\usepackage{microtype}      
\usepackage{fancyhdr}       
\usepackage{graphicx}       
\usepackage{listings}       
\usepackage{xcolor}         
\usepackage{float}          
\graphicspath{{fig/}}       

\title{Integrating External Tools with Large Language Models (LLM) to Improve Accuracy}

\author{
  Nripesh Niketan\textsuperscript{1}\thanks{ORCID: 0009-0008-2066-1937} \\
  \texttt{nripesh14@gmail.com} \\
   \And
  Hadj Batatia\textsuperscript{1}\thanks{ORCID: 0000-0003-0433-2152} \\
  \texttt{h.batatia@hw.ac.uk} \\
}

\begin{document}
\thispagestyle{empty} 
\maketitle

\begin{abstract}
This paper deals with improving querying large language models (LLMs). It is well-known that without relevant contextual information, LLMs can provide poor quality responses or tend to hallucinate. Several initiatives have proposed integrating LLMs with external tools to provide them with up-to-date data to improve accuracy. In this paper, we propose a framework to integrate external tools to enhance the capabilities of LLMs in answering queries in educational settings. Precisely, we develop a framework that allows accessing external APIs to request additional relevant information. Integrated tools can also provide computational capabilities such as calculators or calendars. The proposed framework has been evaluated using datasets from the Multi-Modal Language Understanding (MMLU) collection. The data consists of questions on mathematical and scientific reasoning. Results compared to state-of-the-art language models show that the proposed approach significantly improves performance. Our Athena framework achieves 83\% accuracy in mathematical reasoning and 88\% in scientific reasoning, substantially outperforming all tested models including GPT-4o, LLaMA-Large, Mistral-Large, Phi-Large, and GPT-3.5, with the best baseline model (LLaMA-Large) achieving only 67\% and 79\% respectively. These promising results open the way to creating complex computing ecosystems around LLMs to make their use more natural to support various tasks and activities.
\end{abstract}


\section{Introduction}

The recent development of Large Language Models (LLMs), such as GPT, BERT~\cite{devlin2018bert}, and LLaMA~\cite{touvron2023llama}, has had a big impact on natural language processing (NLP) and artificial intelligence (AI). These models have shown an impressive ability to comprehend and produce human-like text and are powered by extensive datasets and complex algorithms. Current LLM models are great in handling and producing natural language but face difficulties with tasks that demand access to current data or active computational capabilities. For example, responding to inquiries about current stock market trends or solving complex mathematical problems is beyond their reach. This limitation is largely due to LLMs being trained on fixed datasets and their limited ability to directly connect with external databases or computational tools.

To overcome these challenges, it is becoming necessary to integrate LLMs with external tools like calculators, calendars, and databases. This combination improves the capabilities of LLMs, allowing them to process language while having access to and analysing current data, and handling computational tasks. This expansion broadens their practical use and application by a large margin. Recent developments in LLMs have focused on extending their capabilities through external tools to address tasks like arithmetic, factual lookups, and real-time information retrieval. Integrating external tools with LLMs methods can be classified into four major categories: Retrieval-augmented generation (RAG), Code execution and computation, connection to APIs, Hybrid systems. Retrieval-augmented methods aim at connecting LLMs with external databases or retrieval systems, such as search engines and databases, to retrieve real-time data in order to provide more accurate, industry-specific, and relevant answers~\cite{gao2023retrieval,lewis2020retrieval}. Integrating code execution and computation tools, like Python, data analysis, solvers, calculator, and symbolic reasoners, allows executing code, performing mathematical computations to enhance LLMs capabilities to solve complex tasks~\cite{austin2021program,chen2021evaluating,welleck2021naturalproofs,cobbe2021training}. Connecting APIs, such as financial, health, weather, to utilise specialised service in order to handle domain-specific tasks~\cite{nakano2021webgpt,komeili2021internet,thoppilan2022lamda,peng2023check}. More general hybrid systems aim at combining symbolic reasoning, knowledge graphs, rule-based engines and other techniques to regularise or guide LLMs towards more deterministic and explainable reasoning~\cite{pan2023unifying,yao2022react,pan2023logic}. In this work, we are interested in frameworks that allow connecting external APIs, such as calculators, calendars, solvers, in order to improve precision of LLM answers in an educational context.

The Toolformer, introduced by Meta AI Research and Universitat Pompeu Fabra, enables LLMs to autonomously use simple APIs. This model employs a self-supervised loss to generate a language modelling dataset with embedded API calls, which is then fine-tuned to enhance future token predictions. Toolformer incorporates various tools like calculators and search engines, demonstrating improved zero-shot performance on downstream tasks and addressing limitations such as fact hallucination and outdated information~\cite{schick2023toolformer}. The Gorilla model, based on a fine-tuned LLaMA model, focuses on enhancing API interaction within LLMs. It surpasses GPT-4 in generating accurate API calls and adapting to document changes, significantly reducing hallucination issues. Gorilla is tested against the APIBench dataset, which includes diverse APIs, showcasing its ability to handle over 1,600 APIs effectively. This model enhances the practical application of LLMs in real-world scenarios by connecting them to a broad spectrum of available APIs~\cite{patil2023gorilla}.

Integrating symbolic reasoning with Large Language Models (LLMs) has also been investigated to enhance their ability to execute arithmetic and other computational tasks, where deterministic solutions are critical. Models such as Program-Aided Language models (PAL) and ToolkenGPT have been pivotal in this development. PAL introduces a novel approach by generating Python programs as intermediate reasoning steps. This method leverages the LLMs' language understanding and the precise execution of a Python interpreter, significantly improving arithmetic and symbolic reasoning tasks. For instance, PAL outperformed traditional models on the gsm8k math problem benchmark by 8\%, and by 40\% on the more challenging gsm-hard version~\cite{gao2022pal}. ToolkenGPT utilizes a unique ``toolken'' token that triggers specific tool usage within the model, enhancing both fine-tuning and in-context learning. This framework allows the LLM to dynamically adapt to an expanding set of tools, showing strong performance in tasks requiring numerical reasoning and knowledge-based question answering~\cite{hao2023toolkengpt}.

Frameworks connecting LLMs with APIs have revolutionized task execution across various domains, with models like TaskMatrix.AI and Gorilla leading the way. TaskMatrix.AI, developed by Microsoft, integrates foundation models with APIs to tackle a wide range of tasks, leveraging collective intelligence from various models. This system combines a Multimodal Conversational Foundation Model for user interaction with a vast API platform to execute tasks effectively across digital and physical spaces~\cite{liang2023taskmatrix}. The Gorilla model enhances LLMs' ability to interact with APIs, bridging the gap between natural language processing and real-world application by facilitating the use of over 1,600 APIs. This model adapts to document changes and reduces hallucinations, improving how users interact with digital tools through natural language queries~\cite{patil2023gorilla}.

The LATM (LLMs As Tool Makers) framework represents a shift in LLM applications from using external tools to creating and utilizing bespoke tools. This approach enhances problem-solving capabilities and reduces dependency on external resources. LATM operates in two phases: tool making and tool using. Initially, LLMs create tools as Python functions tailored for specific tasks. These tools are then used by the same or different LLMs for problem-solving, allowing for a flexible, cost-effective approach. This framework has been validated in tasks like the Big-Bench, showing that it can match higher-cost models in performance while reducing inference costs. LATM demonstrates a practical, scalable method for enhancing LLMs' functionality, potentially transforming their role in AI by enabling them to independently create and apply tools for complex tasks.

This paper focuses on creating a framework that supports the integration of Large Language Models (LLMs) with external tools. The study covers tools and technologies to implement the integration of LLMs with external tools and evaluates its effectiveness in real-world applications.

\section{Proposed Framework}

In order to allow LLMs to use external tools to enhance their capabilities, we designed a framework, named Athena. The system manages API of external tools that can be used to enable the LLM to provide accurate, up-to-date, data-driven responses across various domains.

\subsection{Architecture}

\begin{figure}[H]
  \centering
  \includegraphics[width=0.8\textwidth]{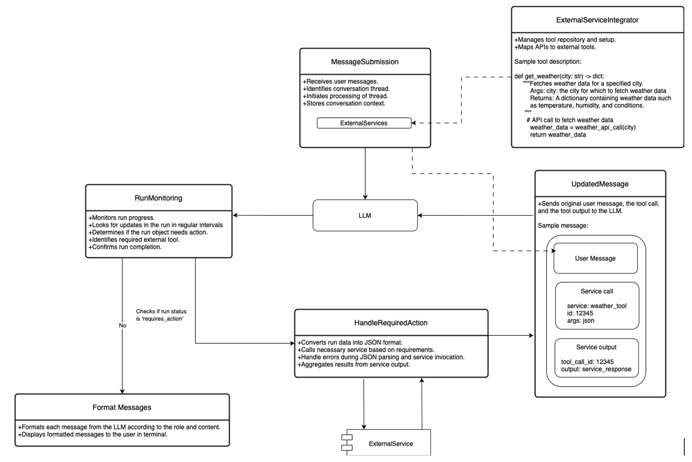}
  \caption{Architecture of the proposed Athena framework showing the integration of external tools with LLMs through various components including ExternalServiceIntegrator, MessageSubmission, RunMonitoring, HandleRequiredAction, and UpdateMessage services.}
  \label{fig:architecture}
\end{figure}

The architecture of the proposed framework is shown in Figure~\ref{fig:architecture}. The ExternalServiceIntegrator component manages the tools repository. Any added tool is provided to the system using a schema-like structure, implemented using frameworks like Pydantic, which specifies information about the tool such as specific functionalities, comprehensive descriptions, and required parameters. More precisely the schema includes (Listing~\ref{lst:tool_format}) the tool's name, description, and the types and details of the arguments it accepts. This structured approach allows the LLM to have explicit knowledge of each tool's capabilities and requirements.

\begin{lstlisting}[caption={Format of a tool description.}, label={lst:tool_format}]
def add(a: int, b: int) -> int:
    """ Adds a and b.
    Args:
        a: first int
        b: second int
    """
    return a + b
\end{lstlisting}

The LLM's awareness of the available tools is facilitated by the registration of these tools within its operational environment, typically achieved through a configuration that includes each tool's detailed schema. These schemas act like blueprints that tell the LLM what each tool does, what inputs it needs, and the kind of response it generates. By referencing these schemas, the LLM can match tools to user queries requiring specific external data or computations that are beyond the LLM's internal processing capabilities.

The MessageSubmission component implements a user interface that allows submitting user queries and managing context.

The RunMonitoring service is responsible for identifying required external tools. The decision-making process regarding when to use these tools starts with the LLM analysing the user's input. This analysis involves parsing the query to extract key information and intent. If the query aligns with the capabilities of one or more registered tools—determined by keyword matching, intent recognition, or query complexity—the LLM then identifies this as an opportunity to use an external tool. For instance, a question about current weather conditions might trigger the identification of a weather data tool based on keywords like ``weather'' and the inclusion of a geographical location.

Once a tool is deemed appropriate for a query, the HandleRequiredAction service proceeds to extract the necessary parameters from the user input. This extraction process uses natural language understanding techniques to identify and map the required data points from the query to the parameters defined in the tool's schema. For example, if the tool requires a city name and date to fetch weather data, the LLM extracts these details from the query.

After parameter extraction, the service formats these parameters according to the API's expected structure, ensuring that all required data points are correctly included. This often involves transforming natural language inputs into more structured data formats that the API can process, such as converting a city's name into a standardized location code.

The system then sends the formatted parameters to the external tool's API. Upon receiving the results from the API, the returned data is integrated into the ongoing dialogue. This integration is handled by converting the API's raw output into a natural, conversational response that aligns with the user's original query and maintains the flow of the dialogue. The UpdateMessage service submits the newly updated query to the LLM.

This entire process is iterative; the LLM continuously assesses whether additional information from external tools is needed to fully address the user's query. This assessment might lead to multiple rounds of tool invocations until the query is completely answered. Finally, the comprehensive response, enriched with both the LLM's internal knowledge and the external tool's specialized data, is delivered to the user.

This streamlined methodology enables the LLM to effectively augment its responses with specialized external data, thereby enhancing the accuracy and relevance of the information provided to the user.

\textbf{Langchain Implementation:} In the LangChain implementation of the Athena framework, the system uses the Unify~\cite{unify2024} platform in conjunction with the LangChain framework. Unify is different from traditional LLMs. It functions as a comprehensive hosting tool that aggregates various open-source LLMs, providing access to them through a unified API. This setup allows users to use a diverse range of LLMs tailored to different tasks and capabilities. The LangChain framework is integral to this implementation. It acts as middleware that integrates the external tools with the LLMs hosted on the Unify platform. In this setup, LangChain abstracts the complexity of tool integration from the user and streamlines the process of increasing LLM capabilities with external APIs.

In practical terms, the LangChain implementation operates under a less hands-on approach from the developers or users in terms of direct API management. Instead of manually preparing and managing API calls, users simply configure LangChain to recognize and utilize the available tools. This part of the system's operation—deciding which LLM to deploy for a given task and how to integrate the response from an external tool—is managed internally by LangChain.

\section{Evaluation}

In order to evaluate the proposed framework, we integrated a few tools and ran various experiments on mathematical and scientific reasoning. This section describes the integrated tools, the experiments and the results.

\subsection{Sample Integrated Tools}

The proposed framework is generic and allows integrating any API. However, in this study, we focused on a few important ones, namely ArXiv, Google SERPer, OpenWeatherMap, Google Calendar, and Wolfram Alpha.

\begin{itemize}
\item As Athena's computational backbone, the Wolfram Alpha API supports complex calculations and algorithm-based queries across various scientific and mathematical fields.
\item The Google SERPer API enables the system to perform web searches and deliver relevant online content in response to user queries. This tool is critical for extending the model's knowledge beyond its training data.
\item ArXiv API allows the system to access and provide detailed information on scholarly articles. It aids users in retrieving and understanding academic content quickly, enhancing research efficiency.
\item The OpenWeatherMap API provides real-time weather forecasts and historical data, allowing the AI to assist users in weather-related planning and inquiries effectively.
\item Google Calendar is integrated to manage scheduling and time-based tasks. This feature allows users to interact with their calendar through natural language commands, supported by secure authentication and event management functionalities.
\end{itemize}

\subsection{Datasets}

Specific datasets from the Multi-Modal Language Understanding (MMLU) collection hosted on HuggingFace were chosen to test the Athena framework against various state-of-the-art language models. These datasets contain multiple-choice questions designed to assess the models' proficiency across different domains and educational levels. The structured format of these datasets, where each entry includes a question, four potential answers, and the correct answer, provides a standardized method to measure the accuracy of the AI models' responses. The selected datasets focus mostly on mathematical and scientific disciplines at various educational stages.

For mathematics, datasets labelled as Elementary Mathematics, High School Mathematics, and College Mathematics were used. To create a comprehensive math testing dataset, 33 questions were selected from the Elementary and High School Mathematics set, and 34 questions from the College Mathematics set. This selection ensures a balanced representation of both basic and advanced mathematical problems. This approach challenges the models to handle a range of complexities and mathematical concepts.

Similarly, for science, the chosen datasets included High School and College levels for Physics, Chemistry, and Biology. High School Physics, Chemistry, Biology contributed 16 questions each; and each of College Physics, Chemistry, Biology contributed 17 questions, with a total of 100 science questions. This selection tests the model's ability to interpret and solve scientific problems that require fact-based knowledge and the application of scientific theories. This diverse range of subjects and difficulty levels in these datasets also allows for a thorough evaluation of how well the model can answer different types of inquiries.

\subsection{Experimental Scenario}

\textbf{Testing Procedure:} A Jupyter notebook was used to systematically test the models on the selected multiple-choice questions from the MMLU datasets. The testing process involved a setup where each question from the datasets was formatted in a specific way for consistency and to simulate a natural questioning environment. Each question was presented to the Large Language Model (LLM) along with the corresponding multiple-choice options. The format was structured to include the question followed by the options, each labelled with letters (A, B, C, D). To standardize the evaluation and to capture the model's responses in a structured manner, the LLM was instructed to output its answers in JSON format. This requirement was specified in the prompt to the model to ensure that the output could be easily parsed and analysed. Listing~\ref{lst:response_format} shows an example of the specific format requested, showing clear separation between the chosen answer and the text of the answer, helping in the automated evaluation of responses.

\begin{lstlisting}[caption={Format of response.}, label={lst:response_format}]
"{question}
Options: {options}
I want you to give me the output in the form of json.
Example:
'''json {
    "answer": "<The right option (A, B, C, D)>",
    "value": "<Value of multiple choice answer>",
} '''"
\end{lstlisting}

\textbf{Recording and Evaluation:} During the test runs, each response from the LLM was recorded from the Jupyter notebook. The responses were then compared to the correct answers provided in the datasets. For each response, it was recorded whether the LLM correctly identified the right option (A, B, C, or D). The primary metrics noted were the number of correct and incorrect answers, which were used to calculate the accuracy of the LLM for both the mathematics and science questions separately.

\textbf{Comparison:} The Athena framework and all baseline language models (GPT-3.5, GPT-4o, LLaMA-Large, Mistral-Large, and Phi-Large) were presented with the exact same set of questions under identical conditions to ensure a fair and controlled comparison. This methodology aimed to provide a clear picture of how each model performs when faced with identical academic challenges, focusing on their ability to interpret and solve mathematical and scientific problems. The comparison tests were designed to highlight the differences in accuracy and capacity across all evaluated models.

\subsection{Results}

This section presents the results of applying the proposed Athena framework to mathematical and scientific reasoning datasets, comparing its performance against state-of-the-art language models including GPT-3.5, GPT-4o, LLaMA-Large, Mistral-Large, and Phi-Large.

\begin{figure}[H]
  \centering
  \includegraphics[width=0.8\textwidth]{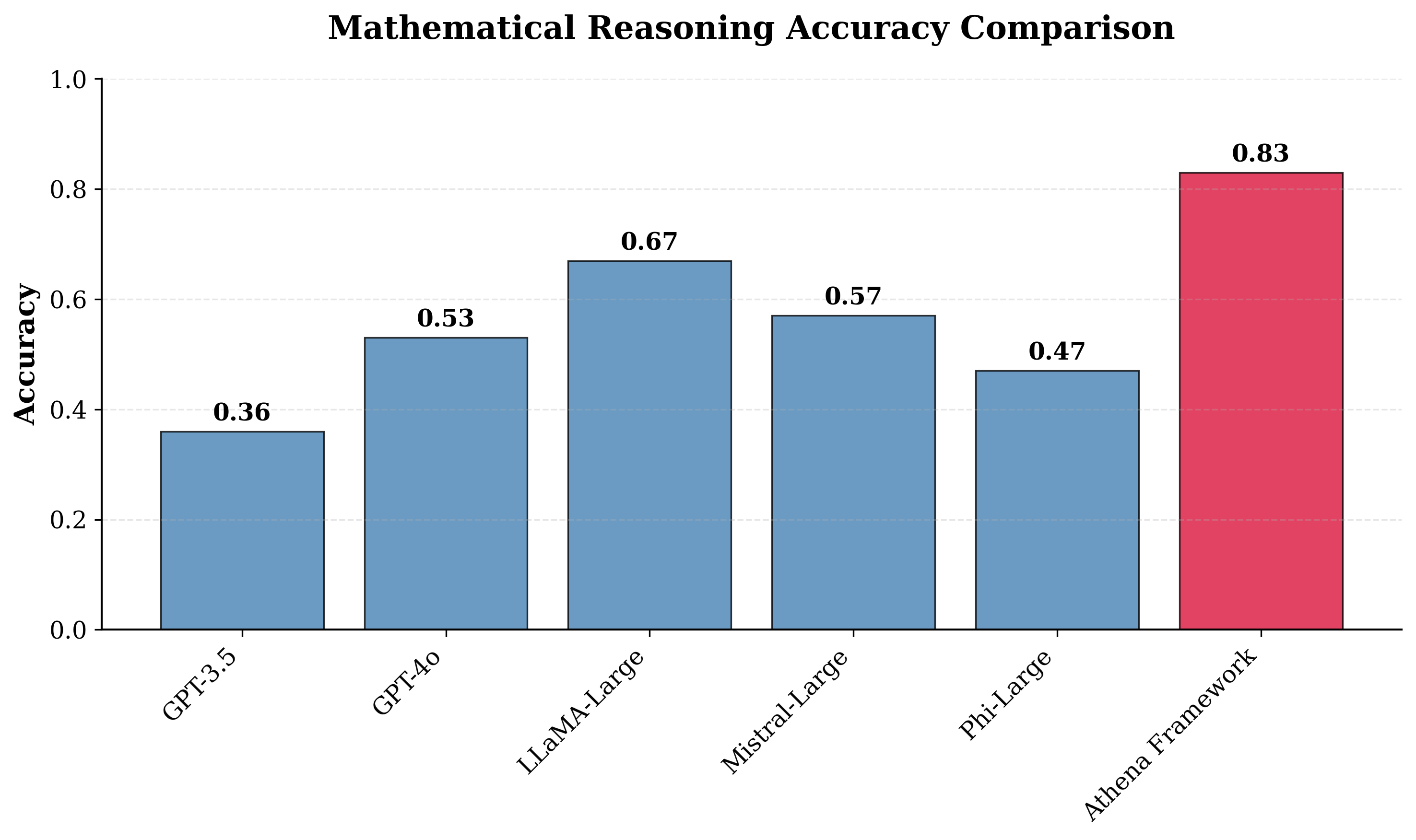}
  \caption{Mathematical reasoning accuracy comparison between Athena framework and state-of-the-art language models using MMLU mathematics dataset.}
  \label{fig:math_results}
\end{figure}

\textbf{Mathematical Reasoning Results:} The evaluation on mathematical questions revealed significant performance differences across models (Figure~\ref{fig:math_results} and Table~\ref{tab:math_comparison}). Among the baseline language models, LLaMA-Large achieved the highest accuracy at 67\%, followed by Mistral-Large at 57\%, GPT-4o at 53\%, Phi-Large at 47\%, and GPT-3.5 at 36\%. These models showed varying capabilities in handling mathematical problems, with performance generally limited to questions that could be solved through memorized knowledge rather than computational reasoning.

\begin{table}[H]
\centering
\caption{Mathematical reasoning accuracy comparison}
\label{tab:math_comparison}
\begin{tabular}{lc}
\toprule
Model & Accuracy \\
\midrule
GPT-3.5 & 0.36 \\
GPT-4o & 0.53 \\
LLaMA-Large & 0.67 \\
Mistral-Large & 0.57 \\
Phi-Large & 0.47 \\
\midrule
\textbf{Athena Framework} & \textbf{0.83} \\
\bottomrule
\end{tabular}
\end{table}

In contrast, the Athena framework achieved an accuracy of 83\%, substantially outperforming all baseline models. This improvement was largely attributed to Athena's ability to leverage integrated computational tools, particularly calculators, for numerical problem-solving. The framework demonstrated strong capacity to handle diverse mathematical challenges, from basic arithmetic to complex algebraic problems that required step-by-step computational reasoning beyond what standalone LLMs could provide.

\begin{figure}[H]
  \centering
  \includegraphics[width=0.8\textwidth]{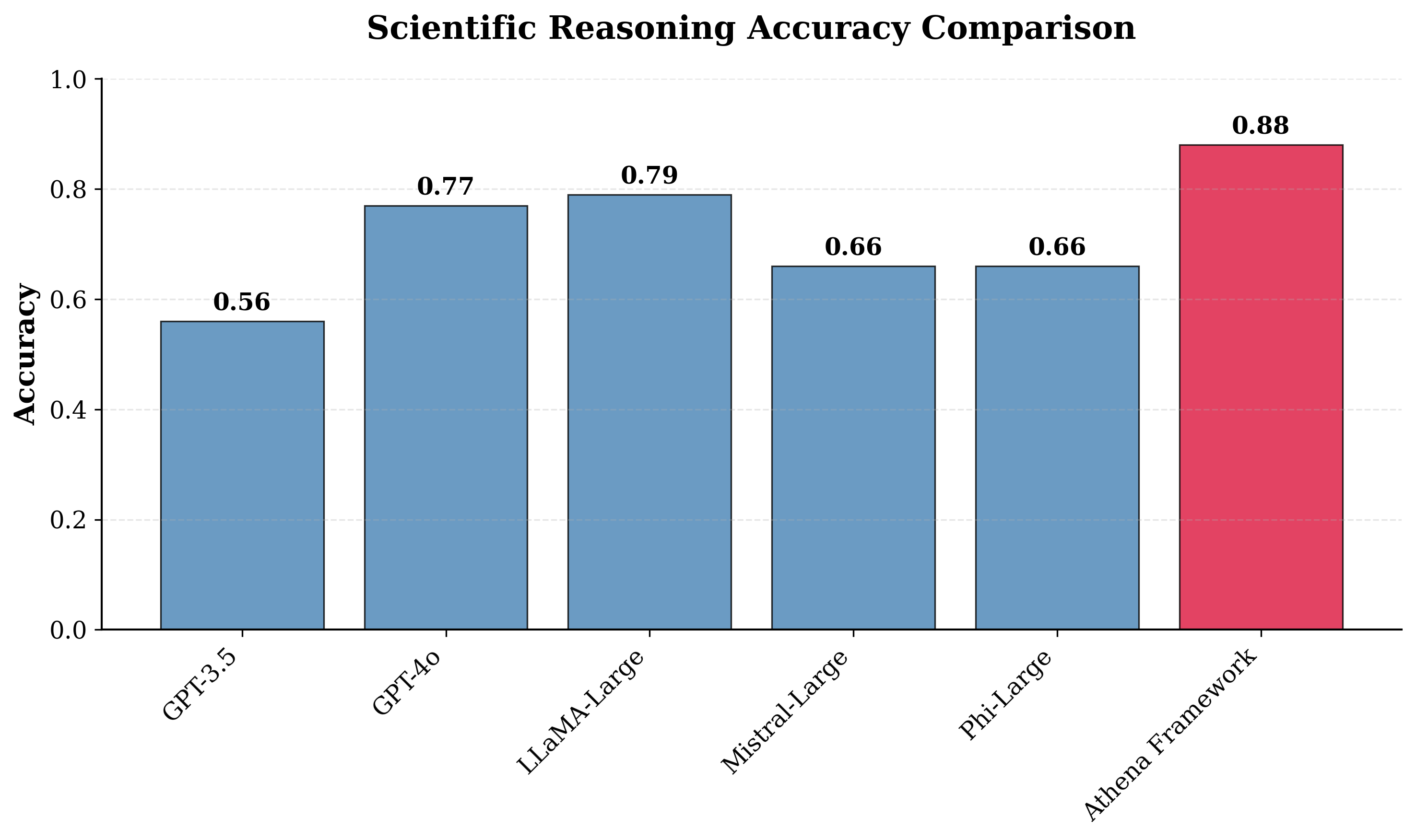}
  \caption{Scientific reasoning accuracy comparison between Athena framework and state-of-the-art language models using MMLU science dataset.}
  \label{fig:science_results}
\end{figure}

\textbf{Scientific Reasoning Results:} The scientific reasoning evaluation showed similar patterns, though with generally higher baseline performance across all models (Figure~\ref{fig:science_results} and Table~\ref{tab:science_comparison}). LLaMA-Large again performed best among baseline models with 79\% accuracy, followed by GPT-4o at 77\%, while Mistral-Large and Phi-Large both achieved 66\%, and GPT-3.5 reached 56\%. The baseline models showed particular strength in theoretical questions involving direct recall of scientific concepts and definitions.

\begin{table}[H]
\centering
\caption{Scientific reasoning accuracy comparison}
\label{tab:science_comparison}
\begin{tabular}{lc}
\toprule
Model & Accuracy \\
\midrule
GPT-3.5 & 0.56 \\
GPT-4o & 0.77 \\
LLaMA-Large & 0.79 \\
Mistral-Large & 0.66 \\
Phi-Large & 0.66 \\
\midrule
\textbf{Athena Framework} & \textbf{0.88} \\
\bottomrule
\end{tabular}
\end{table}

The Athena framework achieved 88\% accuracy in scientific reasoning, maintaining its superior performance across domains. Out of 100 science questions, Athena incorrectly answered only 12, demonstrating substantial capability to handle a broad spectrum of scientific inquiries. The framework's success was particularly evident in questions requiring numerical calculations combined with theoretical knowledge, showcasing its ability to effectively integrate computational tools with factual data retrieval.

\textbf{Analysis and Implications:} The results demonstrate that while modern LLMs have improved significantly over earlier generations, tool integration provides capabilities that cannot be achieved through model scaling alone. The performance gap between Athena and the best baseline model (LLaMA-Large) was 16 percentage points in mathematics and 9 percentage points in science. The smaller gap in scientific reasoning compared to mathematical reasoning suggests that newer models have improved significantly in handling factual scientific knowledge, but still benefit substantially from computational tools for calculation-intensive problems.

These findings confirm that tool integration remains a valuable approach even as base model capabilities improve, and that the Athena framework provides consistent benefits across different types of reasoning tasks. The framework's ability to leverage external computational resources enables it to handle complex problems that require both linguistic understanding and precise mathematical computation.

\section{Conclusion}

This paper presented the Athena framework for integrating external tools with LLMs to enhance the accuracy of model response. The framework allows for the integration of any tool. A study was conducted using a set of typical tools that can benefit educational applications. The evaluation using mathematics and science questions from the MMLU dataset demonstrated significant improvements over standalone LLM performance.

The key contributions of this work include: (1) A flexible framework architecture that allows seamless integration of external APIs and computational tools with LLMs, (2) Comprehensive evaluation demonstrating significant performance improvements in educational domains, with 83\% accuracy in mathematics and 88\% in science, substantially outperforming all tested state-of-the-art models including GPT-4o, LLaMA-Large, Mistral-Large, and Phi-Large, (3) Evidence that tool integration provides capabilities that cannot be achieved through model scaling alone, and (4) A practical implementation using LangChain and Unify platforms that abstracts the complexity of tool integration.

The results indicate that while modern LLMs have improved substantially, there remains a significant advantage in augmenting them with specialized external tools, particularly for tasks requiring computational capabilities or access to current information. Future work will focus on expanding the range of integrated tools, improving the decision-making process for tool selection, and exploring applications in other domains beyond education.

\section*{Acknowledgments}

This work was supported by Heriot-Watt University. We thank the reviewers for their valuable feedback and suggestions that helped improve this paper.

\bibliographystyle{unsrt}  
\bibliography{references}  

\begin{thebibliography}{10}

\bibitem{devlin2018bert}
Jacob Devlin, Ming-Wei Chang, Kenton Lee, and Kristina Toutanova.
\newblock Bert: Pre-training of deep bidirectional transformers for language understanding.
\newblock {\em arXiv preprint arXiv:1810.04805}, 2018.

\bibitem{touvron2023llama}
Hugo Touvron, Thibaut Lavril, Gautier Izacard, Xavier Martinet, Marie-Anne Lachaux, Timoth{\'e}e Lacroix, Baptiste Rozi{\`e}re, Naman Goyal, Eric Hambro, Faisal Azhar, et~al.
\newblock Llama: Open and efficient foundation language models.
\newblock {\em arXiv preprint arXiv:2302.13971}, 2023.

\bibitem{gao2023retrieval}
Luyu Gao, Zhuyun Dai, and Jamie Callan.
\newblock Retrieval-augmented generation for knowledge-intensive nlp tasks.
\newblock {\em Advances in Neural Information Processing Systems}, 36, 2023.

\bibitem{lewis2020retrieval}
Patrick Lewis, Ethan Perez, Aleksandra Piktus, Fabio Petroni, Vladimir Karpukhin, Naman Goyal, Heinrich K{\"u}ttler, Mike Lewis, Wen-tau Yih, Tim Rockt{\"a}schel, et~al.
\newblock Retrieval-augmented generation for knowledge-intensive nlp tasks.
\newblock {\em Advances in neural information processing systems}, 33:9459--9474, 2020.

\bibitem{austin2021program}
Jacob Austin, Augustus Odena, Maxwell Nye, Maarten Bosma, Henryk Michalewski, David Dohan, Ellen Jiang, Carrie Cai, Michael Terry, Quoc Le, et~al.
\newblock Program synthesis with large language models.
\newblock {\em arXiv preprint arXiv:2108.07732}, 2021.

\bibitem{chen2021evaluating}
Mark Chen, Jerry Tworek, Heewoo Jun, Qiming Yuan, Henrique Ponde de~Oliveira Pinto, Jared Kaplan, Harri Edwards, Yuri Burda, Nicholas Joseph, Greg Brockman, et~al.
\newblock Evaluating large language models trained on code.
\newblock {\em arXiv preprint arXiv:2107.03374}, 2021.

\bibitem{welleck2021naturalproofs}
Sean Welleck, Jiacheng Liu, Ronan~Le Bras, Hannaneh Hajishirzi, Yejin Choi, and Kyunghyun Cho.
\newblock Naturalproofs: Mathematical theorem proving in natural language.
\newblock {\em arXiv preprint arXiv:2104.01112}, 2021.

\bibitem{cobbe2021training}
Karl Cobbe, Vineet Kosaraju, Mohammad Bavarian, Mark Chen, Heewoo Jun, Lukasz Kaiser, Matthias Plappert, Jerry Tworek, Jacob Hilton, Reiichiro Nakano, et~al.
\newblock Training verifiers to solve math word problems.
\newblock {\em arXiv preprint arXiv:2110.14168}, 2021.

\bibitem{nakano2021webgpt}
Reiichiro Nakano, Jacob Hilton, Suchir Balaji, Jeff Wu, Long Ouyang, Christina Kim, Christopher Hesse, Shantanu Jain, Vineet Kosaraju, William Saunders, et~al.
\newblock Webgpt: Browser-assisted question-answering with human feedback.
\newblock {\em arXiv preprint arXiv:2112.09332}, 2021.

\bibitem{komeili2021internet}
Mojtaba Komeili, Kurt Shuster, and Jason Weston.
\newblock Internet-augmented dialogue generation.
\newblock {\em International Conference on Machine Learning}, pages 8460--8478, 2021.

\bibitem{thoppilan2022lamda}
Romal Thoppilan, Daniel De~Freitas, Jamie Hall, Noam Shazeer, Apoorv Kulshreshtha, Heng-Tze Cheng, Alicia Jin, Taylor Bos, Leslie Baker, Yu~Du, et~al.
\newblock Lamda: Language models for dialog applications.
\newblock {\em arXiv preprint arXiv:2201.08239}, 2022.

\bibitem{peng2023check}
Baolin Peng, Michel Galley, Pengcheng He, Hao Cheng, Yujia Xie, Yu~Hu, Qiuyuan Huang, Lars Liden, Zhou Yu, Weizhu Chen, et~al.
\newblock Check your facts and try again: Improving large language models with external knowledge and automated feedback.
\newblock {\em arXiv preprint arXiv:2302.12813}, 2023.

\bibitem{pan2023unifying}
Shirui Pan, Linhao Luo, Yufei Wang, Chen Chen, Jiapu Wang, and Xindong Wu.
\newblock Unifying large language models and knowledge graphs: A roadmap.
\newblock {\em IEEE Transactions on Knowledge and Data Engineering}, 2023.

\bibitem{yao2022react}
Shunyu Yao, Jeffrey Zhao, Dian Yu, Nan Du, Izhak Shafran, Karthik Narasimhan, and Yuan Cao.
\newblock React: Synergizing reasoning and acting in language models.
\newblock {\em arXiv preprint arXiv:2210.03629}, 2022.

\bibitem{pan2023logic}
Liangming Pan, Alon Albalak, Xinyi Wang, and William~Yang Wang.
\newblock Logic-lm: Empowering large language models with symbolic solvers for faithful logical reasoning.
\newblock {\em arXiv preprint arXiv:2305.12295}, 2023.

\bibitem{schick2023toolformer}
Timo Schick, Jane Dwivedi-Yu, Roberto Dess{\`\i}, Roberta Raileanu, Maria Lomeli, Luke Zettlemoyer, Nicola Cancedda, and Thomas Scialom.
\newblock Toolformer: Language models can teach themselves to use tools.
\newblock {\em arXiv preprint arXiv:2302.04761}, 2023.

\bibitem{patil2023gorilla}
Shishir~G Patil, Tianjun Zhang, Xin Wang, and Joseph~E Gonzalez.
\newblock Gorilla: Large language model connected with massive apis.
\newblock {\em arXiv preprint arXiv:2305.15334}, 2023.

\bibitem{gao2022pal}
Luyu Gao, Aman Madaan, Shuyan Zhou, Uri Alon, Pengfei Liu, Yiming Yang, Jamie Callan, and Graham Neubig.
\newblock Pal: Program-aided language models.
\newblock {\em arXiv preprint arXiv:2211.10435}, 2022.

\bibitem{hao2023toolkengpt}
Shibo Hao, Tianyang Liu, Zhen Wang, and Zhiting Hu.
\newblock Toolkengpt: Augmenting frozen language models with massive tools via tool embeddings.
\newblock {\em arXiv preprint arXiv:2305.11554}, 2023.

\bibitem{liang2023taskmatrix}
Yaobo Liang, Chenfei Wu, Ting Song, Wenshan Wu, Yan Xia, Yu~Liu, Yang Ou, Shuai Lu, Lei Ji, Shaoguang Mao, et~al.
\newblock Taskmatrix.ai: Completing tasks by connecting foundation models with millions of apis.
\newblock {\em arXiv preprint arXiv:2303.16434}, 2023.

\bibitem{unify2024}
Unify.
\newblock Unify: The complete llm platform, 2024.
\newblock Accessed: 2024-01-15.

\end{thebibliography}

\end{document}